\pdfoutput=1

\documentclass[11pt]{article}

\usepackage[final]{acl}

\usepackage{times}
\usepackage{latexsym}

\usepackage[T1]{fontenc}

\usepackage[utf8]{inputenc}

\usepackage{microtype}

\usepackage{inconsolata}

\usepackage{graphicx}
\usepackage{subcaption}
\usepackage{amsmath}  
\usepackage{amssymb}  
\usepackage{mathtools} 
\usepackage{bm}       
\usepackage{xspace}
\usepackage{booktabs}
\usepackage{pifont} 
\newcommand{\cmark}{\ding{51}}%
\newcommand{\xmark}{\ding{55}}%

\newcommand{\eg}{e.g.,\xspace}
\newcommand{\ie}{i.e.,\xspace}
\def\figref#1{Figure~\ref{fig:#1}}

\def\tabref#1{Table~\ref{tab:#1}}

\def\eqref#1{Eq.~\ref{eqn:#1}}

\def\Secref#1{Section~\ref{sec:#1}}
\def\Appref#1{Appendix~\ref{sec:#1}}
\def\seclabel#1{\label{sec:#1}\label{p:#1}}
\def\applabel#1{\label{sec:#1}\label{p:#1}}
\def\secref#1{\S\ref{sec:#1}}

\newcommand{\prompt}[1]{
\vspace{2mm}
\noindent
\fbox{
\begin{minipage}{0.99\linewidth}
\vspace{6pt}
\scriptsize	
#1
\vspace{6pt}
\end{minipage}
}
\vspace{2mm}
}

\title{How Programming Concepts and Neurons Are Shared in \\ Code Language
Models}

\author{
    Amir Hossein Kargaran$^{1}$ \quad
    Yihong Liu$^{1}$ \quad
    François Yvon$^{2}$ \quad
    Hinrich Schütze$^{1}$ \\
    \\
    $^1$LMU Munich \protect\& Munich Center for Machine Learning \\
    $^2$Sorbonne Université \protect\& CNRS, ISIR \\
    \texttt{\{amir, yihong\}@cis.lmu.de}
}

\begin{document}
\maketitle
\begin{abstract}

Several studies have explored the mechanisms of large language models (LLMs) in coding tasks, but most have focused on programming languages (PLs) in a monolingual setting. In this paper, we investigate the relationship between multiple PLs and English in the concept space of LLMs. We perform a few-shot translation task on 21 PL pairs using two Llama-based models. By decoding the embeddings of intermediate layers during this task, we observe that the concept space is closer to English (including PL keywords) and assigns high probabilities to English tokens in the second half of the intermediate layers.
We analyze neuron activations for 11 PLs and English,
finding that while language-specific neurons are primarily
concentrated in the bottom layers, those exclusive to each
PL tend to appear in the top layers. For PLs that are highly
aligned with multiple other PLs, identifying
language-specific neurons is not feasible. These PLs also
tend to have a larger keyword set than other PLs and are
closer to the model's concept space regardless of the
input/output PL in the translation task. Our findings provide insights into how LLMs internally represent PLs, revealing structural patterns in the model's concept space.
Code is available at \href{https://github.com/cisnlp/code-specific-neurons}{\path{https://github.com/cisnlp/code-specific-neurons}}.  

\end{abstract}

\section{Introduction}
Most state-of-the-art autoregressive large language models
(LLMs) perform well on coding
tasks~\citep{chen2021evaluating, hou2024large,
lyu2024automatic, zhu2024deepseek, jiang2024survey}. Including code in the
pre-training data has become a common practice in LLM
pre-training, even for models not specifically designed for
coding~\citep{aryabumi2024code}. Most of these LLMs involved
in coding tasks are pre-trained on multiple programming
languages (PLs)~\citep{li2023starcoder,
zhu2024deepseek, jiang2024survey, guo2024deepseek}.
This raises an intriguing question: How does pre-training on multiple PLs and English affect the behavior of the models' ``concept space''
in coding tasks? More specifically: \textbf{RQ1.} Does the model use English or a PL as a kind of ``pivot'' language? \textbf{RQ2.} Can we identify language-specific neurons for PLs and English? Do PLs
and English influence one another and neurons are shared across PLs and English?

As summarized in \tabref{banner}, we observe both similarities and differences in how LLMs represent natural languages versus PLs.

\begin{table}[t]
    \centering
    \scriptsize
    \resizebox{\columnwidth}{!}{ 
    \begin{tabular}{lcc}
        \toprule
        \textbf{Observation} & \textbf{NLs} & \textbf{PLs} \\        \midrule
        1) English detour & \cmark & \cmark (shared with PLs) \\
        2) High alignment & \cmark (English) & \cmark (other PLs, \eg C\#) \\
        3) English neuron ID & \xmark & \cmark \\
        4) Non-English/PL neuron ID & \cmark &  ? (inconsistent) \\        \bottomrule
    \end{tabular}
    }
    \caption{Differences between natural languages (NLs) and
      programming languages (PLs) in English-centric
      LLMs. 1) LLMs' layers reach non-English tokens through a detour via English~\citep{wendler2024llamas}. The same occurs when outputting PLs, though English is shared with PL tokens (\secref{res-logitlens}). 
    2) Non-English languages show high cross-lingual
    alignment with English in LLMs' intermediate
    layers~\citep{kargaran2024mexa}, while PLs, including
    C\#, exhibit high alignment with each other
    (\secref{res-mexa}). 3), 4) It is challenging to
    identify English-specific neurons whose ablation does
    not  affect non-English. It is easy to find neurons
    specific for Non-English (e.g.,
    French)~\citep{tang-etal-2024-language}. For PLs,
there are    English ablatable neurons  with minimal
    performance degradation over PLs, but for some PLs
(e.g., C\#, see \secref{res-lape}) finding ablatable neurons
    without affecting other PLs is hard.
    }
    \label{tab:banner}
\end{table}

\begin{figure}[t]
    \centering

\includegraphics[width=\linewidth]{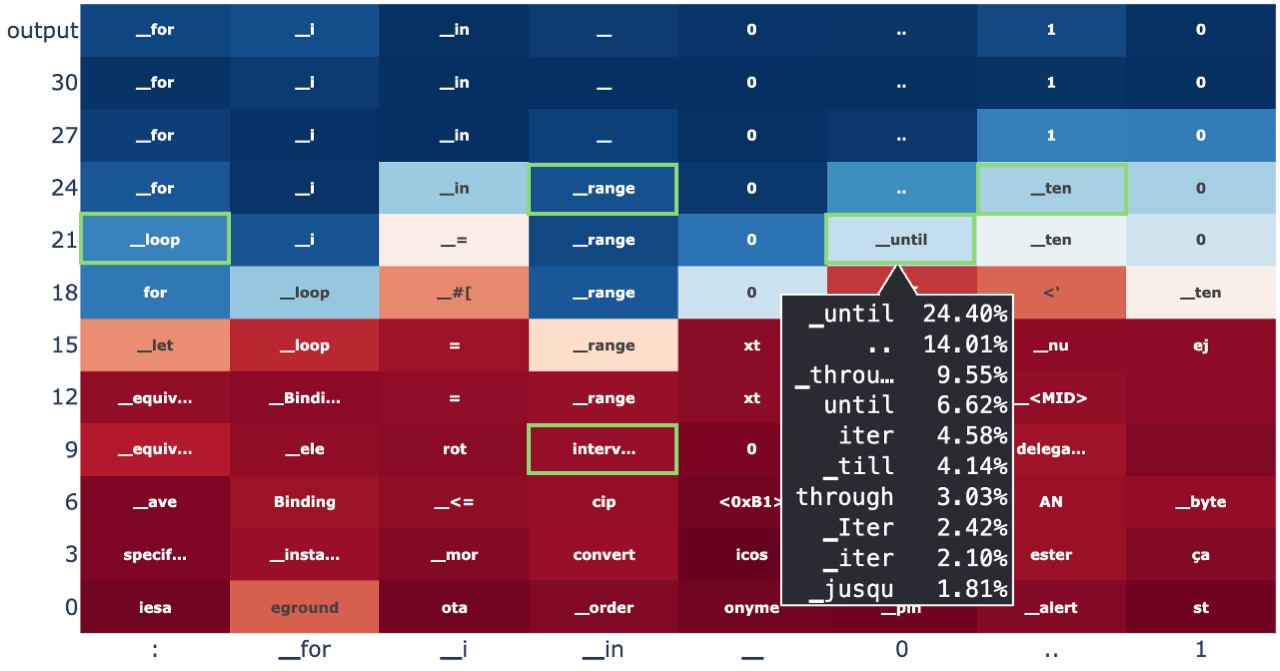}
    \caption{Illustration of logit
      lens~\citep{nostalgebraist2020interpreting} applied to
      CodeLlama 7B for the task of translating a \texttt{for} loop from Java to Rust (showing only Rust loop here). The y-axis shows layers, the x-axis input tokens, and color next-token probabilities (red: low, blue: high). Terms decoded in intermediate layers, such as \texttt{interval}, \texttt{range}, \texttt{until}, and \texttt{ten}, are not keywords in Java or Rust but belong to other PLs (Python, Go, Ruby) and English. In Rust, the \texttt{..} syntax defines a range. The tokens \texttt{until} and \texttt{through}, which decode for the same position but with lower probabilities or in earlier layers, share similar semantics with this syntax.}
    \label{fig:banner}
\end{figure}

\textbf{Contributions.} To investigate the relationship between
English and multiple PLs in the LLM's concept space, we apply methods from the field of interpretability.
Specifically, we focus on two models from the Llama family: CodeLlama 7B~\citep{roziere2023code} and Llama 3.1 8B~\citep{dubey2024llama}. We use the logit lens technique~\citep{nostalgebraist2020interpreting}, which involves applying the ``unembedding'' operation prematurely at intermediate, non-final layers. This approach provides insight into the model’s internal numerical representations, which are otherwise difficult to interpret. We create a super-parallel dataset of 42 translation directions (21 pairs) across seven PLs using the dump of GeeksforGeeks \citep{GeeksforGeeks} prepared by \citet{zhu2022multilingual}. Our results show that the selected Llama models assign high probabilities and top ranks to expected tokens during translation in the last layer, meaning they completely understand the translation task. Tracking token probabilities across layers for different PLs and English using logit lens (see \figref{banner}), we observed: (1) Most tokens in the first half of the layers have low probabilities, near zero, across PLs and English.
(2) Tokens from English and various PLs appear in the intermediate layers, mostly in the second half of the layers.
(3) Most tokens belong to English, followed by all PLs. Among individual PLs, the output PL comes next, followed by PLs such as C++ and C\#, which have some of the largest keyword sets. We use our super-parallel data and measure the cross-lingual alignment for these PLs and find that C\# is more aligned with most languages but not in all cases. For example, the best-aligned PL for PHP and
Python is JavaScript.

We also explore how neuron activations are shared across 11
PLs and English. We use language activation probability
entropy~\citep{tang-etal-2024-language} to identify
language-specific neurons. Our analysis reveals the
following insights: (1) Language-specific neurons are more
concentrated in the bottom layers, followed by another
smaller peak observed around the three quarter point of the layers. (2) Among language-specific neurons, those exclusive to a single PL tend to appear in the top layers.
(3) For PLs such as C\# and Java, which closely align with multiple other PLs, identifying language-specific neurons is challenging.

\section{Materials and methods}
We use three established methods to uncover the concept space of LLMs, using datasets from PLs at parallel, keyword, and raw levels.

\subsection{Datasets}~\seclabel{datasets}
\textbf{Super-parallel PL.}
Most of the parallel datasets available in code community
research (\citet{zhu2022xlcost, zhu2022multilingual,
lachaux2020unsupervised}, \textit{inter alia}) come from
GeeksForGeeks~\citep{GeeksforGeeks}, a website that hosts
data structure and algorithm problems along with solutions
in up to  seven  different PLs: C++, Java, Python, C\#,
JavaScript, PHP, and C. According to GeeksForGeeks, the
solution programs for the same problem follow the same
structure, including consistent variable names, resulting in
semantic consistency across the different languages. In most
cases, the programs for the same problem share the same set
of comments in the same order, indicating that they are
parallel at the snippet level. We use
the \citep{zhu2022multilingual} dump of GeeksForGeeks and
create
a super-parallel dataset for all  seven  PLs, containing 581
parallel code snippets, each available for all  seven  PLs. We
retain only programs that are available in all PLs and that
have the same number of snippets to ensure alignment across
all  seven  PLs.

\textbf{English and PL keywords.} 
We gather, for 22 PLs, programming-specific keywords, as
well as the names of other built-ins
starting
from \citep{mcCulloch2022keywords}. For brevity, we refer to
these as PL keywords. We also extract English
keywords from PanLex~\citep{kamholz-etal-2014-panlex}, which
contains words from several thousand languages, including
English. We only keep keywords that the model's tokenizer represents as a single token and remove numbers from this list (if represented numerically).
Note that PLs have a limited vocabulary consisting primarily of keywords whereas natural languages have an extensive and continuously evolving lexicon. Additionally, many PLs are influenced by older PLs~\citep{sebesta2016concepts}, leading to shared structures and common keywords like \texttt{if}, \texttt{for}, \texttt{while}, and \texttt{return}.

\textbf{Raw PL and English.} We take raw code of eleven popular PLs from the GitHub Code dataset~\citep{codeparrot_github_code_2022}. It consists of 115 million code files from GitHub in 32 PLs.
We select the following eleven popular PLs~\citep{githut2024}:  C, C++, C\#, Go, HTML, Java, JavaScript, PHP, Python, Ruby, and Rust. We also use the English Wikipedia as the source for English texts. We limit each language to 50,000 code files/articles.

\subsection{Models}
We focus on models from the Llama family, which are
autoregressive and decoder-only
transformers~\citep{vaswani2017attention}. We focus on
pre-trained models rather than fine-tuned ones with
instruction tuning or RLHF to minimize confounding
factors. We select two models: CodeLlama
7B~\citep{roziere2023code} and Llama 3.1
8B~\citep{dubey2024llama}. We choose models around 7B
parameters, which are considered a base size in the LLM
community. CodeLlama 7B is pre-trained on 500B tokens from a
code-heavy dataset. It is first initialized with Llama
2~\citep{touvron2023llama2} model weights, which are
pre-trained on general-purpose text and code. CodeLlama
is the only family of Llama models introduced primarily for
coding tasks. The latest models in the Llama family are
general foundation models. Llama 3.1 8B is the latest model
in the family with a base size around 7B parameters and is pre-trained on 15
trillion tokens of general-purpose text and code.

\subsection{Method 1: Interpreting latent embeddings}

Following \citet{wendler2024llamas}, we use logit
lens \citep{nostalgebraist2020interpreting} instead of
tuned lens~\citep{belrose2023eliciting} to decode
intermediate embeddings, as tuned lens is trained to map
internal states to the final next-token prediction, which
may lose the signal of interest.  We use logit lens to
find which of the PLs or English is closer to the abstract
concept space of the selected models.

\textbf{Logit lens.} A transformer model at layer $\ell$ can
be viewed in two parts: (i) a lower part, which includes
layers up to and including layer $\ell$, that maps input
tokens to hidden states, and (ii) an upper part, which
includes layers after $\ell$ that convert hidden states into
logits. The core idea of logit lens is to see the lower part
as a complete transformer and apply $\mathbf{W_U}$, the
``unembedding'' matrix, to project the hidden state at
layer $\ell$, $\mathbf{h}^{(\ell)}$, into logit
scores. These logit scores are then transformed into token
probabilities via the softmax operation. The logit lens operation can be defined as:
\[
\mathrm{LogitLens}(\mathbf{h}^{(\ell)}) = \mathrm{LayerNorm}[\mathbf{h}^{(\ell)}] \mathbf{W_U}.
\]

\textbf{Few-shot translation.} The task is to translate the preceding PL (e.g., Java) code snippet into another PL (e.g., Python). We show the model four code snippets  with their correct translations, followed by a fifth code snippet without its translation, and ask the model to predict the next tokens. With such a prompt, the model can infer that it should translate the fifth code snippet. Since the fifth predicted code snippet could diverge at some point and affect all the subsequent tokens, we predict the tokens one by one and replace the previous tokens with the expected ones. We use our super-parallel PL dataset (\secref{datasets}) for the fifth code snippet (both input and output PLs).
For every input token, at each layer, we compute the
probabilities of the top \(\alpha = 10\) decoded tokens using logit lens and
classify them as belonging to English or one or more PLs using the keywords dataset (\secref{datasets}).

As for the four-shot code snippets, we always use parallel data for basic structures, as shown in the example below (Input PL: Java, Output PL: Python).

\prompt{
\texttt{Java: String message = ""; - Python: message = ""\\
Java: public class MyClass \{\} - Python: class MyClass:\\
Java: public int value = 5; - Python: value = 5\\
Java: public void doSomething() \{\} - Python: def do\_something():\\
Java: for (int i = 0; i < 10; i++) - Python: 
}}

\subsection{Method 2: Cross-lingual alignment}
We employ MEXA~\citep{kargaran2024mexa}, a measure of cross-lingual alignment, to determine which PL aligns most closely with the majority of the selected PLs in the model's intermediate layers. To compute MEXA, we generate code snippet embeddings using position-weighted averaging~\citep{muennighoff2022sgpt} and assess alignment based on cosine similarity comparisons. The higher the score, the greater the alignment, with values ranging between 0 and 1.

\textbf{MEXA.} Given a decoder-only transformer model $m$, MEXA computes the cross-lingual alignment score for language $L_1$ relative to a pivot language $L_2$. Let $S = \{s_1, s_2, \dots, s_n\}$ be a set of $n$ parallel sentences (\ie code snippets) in $L_1$ and $L_2$.  We use our super-parallel  dataset (\secref{datasets}) for each pair. First, we compute sentence embeddings using model \( m \) at layer \( \ell \) with position-weighted averaging. Given a sentence \( s \), its corresponding embedding is denoted as \( \mathbf{e}^{(\ell)}(s) \).
%
We construct a similarity matrix $\mathbf{C}(L_1, L_2, m, \ell) \in \mathbb{R}^{n \times n}$, where each element $c_{ij}(\ell)$ represents the cosine similarity between the embeddings of sentence $s_i$ in $L_1$ and sentence $s_j$ in $L_2$.
%
%
The diagonal elements $c_{ii}(\ell)$ correspond to the similarity between parallel sentence pairs. The MEXA alignment score for matrix $\mathbf{C}(L_1, L_2, m, \ell)$ is defined as:
\[
    \frac{1}{n} \sum_{i=1}^{n} \mathbb{I} \left( c_{ii}(\ell) > \max_{j \neq i} \{ c_{ij}(\ell), c_{ji}(\ell) \} \right),
\]
where $\mathbb{I}$ is the indicator function, which returns 1 if the condition holds and 0 otherwise. This measures how often a parallel sentence pair has the highest similarity compared to any non-parallel pairs.

\subsection{Method 3: Language-specific neurons}

We use language activation probability entropy (LAPE) \citep{tang-etal-2024-language}, which outperforms similar methods in identifying language-specific regions across natural languages. We use LAPE to identify language-specific neurons in each model and analyze their impact on other languages.

\textbf{Neurons in FFN.} Llama-based models~\citep{touvron2023llama1} use a transformer architecture with a GLU variant~\citep{shazeer2020glu}. Like other transformer architectures, their core building blocks include multi-head self-attention (MHA) and feed-forward networks (FFNs). Let $\tilde{\mathbf{h}}^{(\ell)}$ denote the output of the MHA module in the $\ell$-th layer, computed using the previous layer's hidden states and trainable parameters. The FFN module, which outputs the hidden state $\mathbf{h}^{(\ell)} \in \mathbb{R}^{d_1}$, in a GLU variant transformer is given by:
\[
    \mathbf{h}^{(\ell)} = \big(\phi(\tilde{\mathbf{h}}^{(\ell)}\mathbf{W}^{(\ell)}_1) \otimes \tilde{\mathbf{h}}^{(\ell)}\mathbf{W}^{(\ell)}_3\big) \cdot \mathbf{W}^{(\ell)}_2,
\]
where
$\mathbf{W}^{(\ell)}_1, \mathbf{W}^{(\ell)}_3 \in \mathbb{R}^{d_1 \times
d_2}$ and $\mathbf{W}^{(\ell)}_2 \in \mathbb{R}^{d_2 \times
d_1}$ are learnable parameters, and $\phi(\cdot)$ denotes
the activation function. In LAPE, a \textit{neuron} is
defined as the linear transformation of a single column in
$\mathbf{W}^{(\ell)}_1$ followed by the application of the
non-linear activation function. Thus, each FFN module contains $d_2$ neurons. A neuron indexed by $r$ in the $\ell$-th FFN layer is considered ``active'' if its activation value $\phi(\tilde{\mathbf{h}}^{(\ell)}\mathbf{W}^{(\ell)}_1)_{r}$ exceeds zero.

\textbf{LAPE.} To compute LAPE, we feed LLMs different texts, each written in a single language from raw PL and English texts (\secref{datasets}). For the $r$-th neuron in the $\ell$-th layer, we calculate the activation probability when processing texts in language $z$:
\[
    p^z_{\ell,r} = \mathbb{E}\left(\mathbb{I}\big(\phi(\tilde{\mathbf{h}}^{(\ell)} \mathbf{W}^{(\ell)}_1)_r > 0\big) \mid \text{language } z \right),
\]
where $\mathbb{I}$ is the indicator function. This probability is empirically estimated as the likelihood that the neuron's activation value exceeds zero. We obtain the probability  
distribution across languages and normalize it via sum
normalization to compute the normalized probability
$p'^z_{\ell,r}$ for each language $z$. The entropy of this
distribution is:
%
\[
    \text{LAPE}_{\ell,r} = -\sum_{z \in \mathcal{L}} p'^z_{\ell,r} \log (p'^z_{\ell,r}).
\]
where $ \mathcal{L} $ is the set of languages.
We designate neurons with low LAPE scores as ``language-specific neurons,'' as they show a predilection for activation in response to one or two languages, while showing reduced activation probabilities for others. A neuron is deemed specific
to language \( z \) if its corresponding activation probability $p^z_{\ell,r}$ surpasses a predefined threshold.

LAPE is highly dependent on hyperparameter thresholds. The first hyperparameter is the activation threshold, set at the activation probability corresponding to the $\tau$ quantile. The default for LAPE is $\tau = 0.95$. For CodeLlama 7B/Llama 3.1 8B, this corresponds to activation probability thresholds of 0.531 and 0.554, respectively, meaning selected neurons must exhibit activation probabilities exceeding these values for at least one language. The second threshold, the filter threshold $\gamma$, retains only a small fraction of neurons as language-specific by selecting those in the bottom $\gamma$ of LAPE scores. The default setting is $\gamma = 0.01$. However, since this results in varying numbers of selected neurons across languages and makes the comparison between different languages harder to interpret, we instead compute the average number of selected neurons and select the same number, $\nu$, for each language. For the default settings of both selected models, $\nu$ is around 400.

\textbf{Controlled generation.}
To assess the impact of the selected neurons, we set their activation values to zero or zero out the corresponding parameters and then measure changes in model performance. Specifically, we compute language model perplexities (PPLs) to examine how much removing these neurons affects various languages.

\begin{figure*}[t]
    \centering
    
    \begin{subfigure}{0.35\textwidth}
        \centering
        \includegraphics[width=\linewidth]{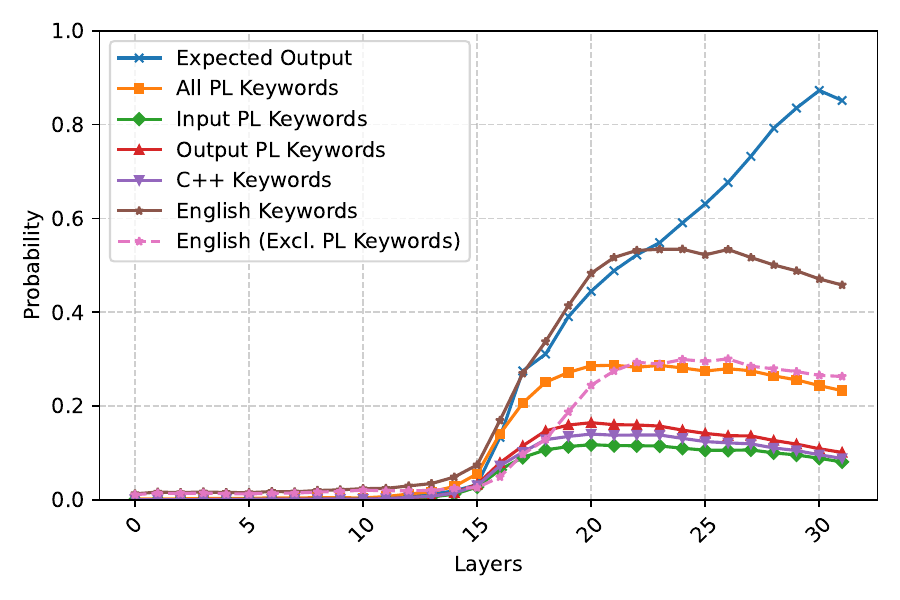}
        \caption{Code Llama 7B; probability value.}
        \label{fig:codellama-prob}
    \end{subfigure}
    \hspace{30pt}
    \begin{subfigure}{0.35\textwidth}
        \centering
        \includegraphics[width=\linewidth]{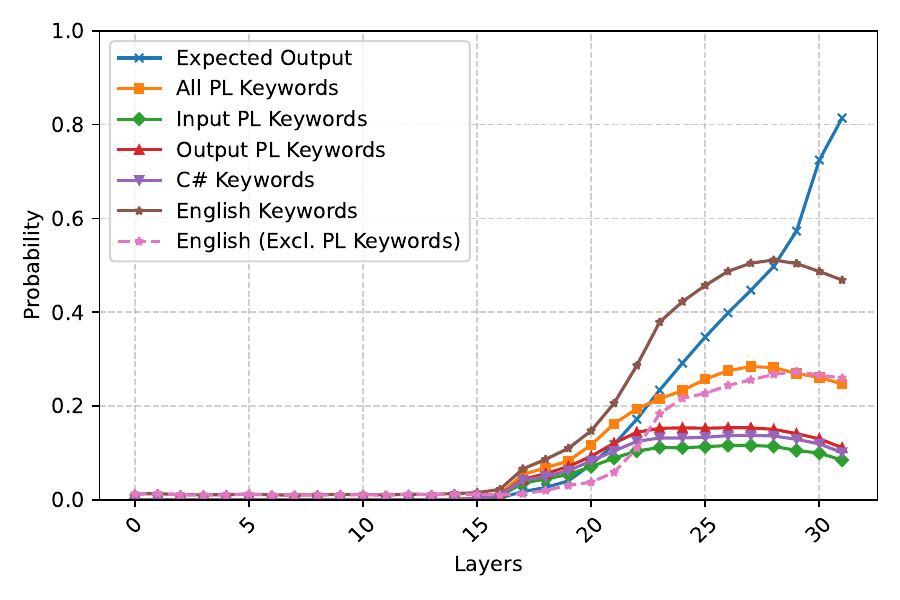}
        \caption{Llama 3.1 8B;  probability value.}
        \label{fig:llama31-prob}
    \end{subfigure}
    
    \begin{subfigure}{0.35\textwidth}
        \centering
        \includegraphics[width=\linewidth]{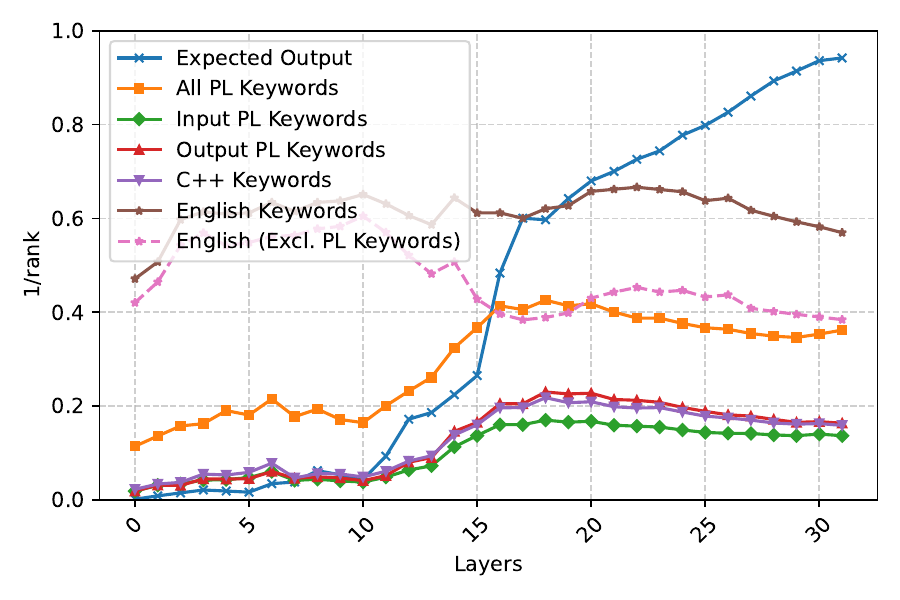}
        \caption{Code Llama 7B; $\frac{1}{rank}$ value.}
        \label{fig:codellama-rank}
    \end{subfigure}
    \hspace{30pt}
    \begin{subfigure}{0.35\textwidth}
        \centering
        \includegraphics[width=\linewidth]{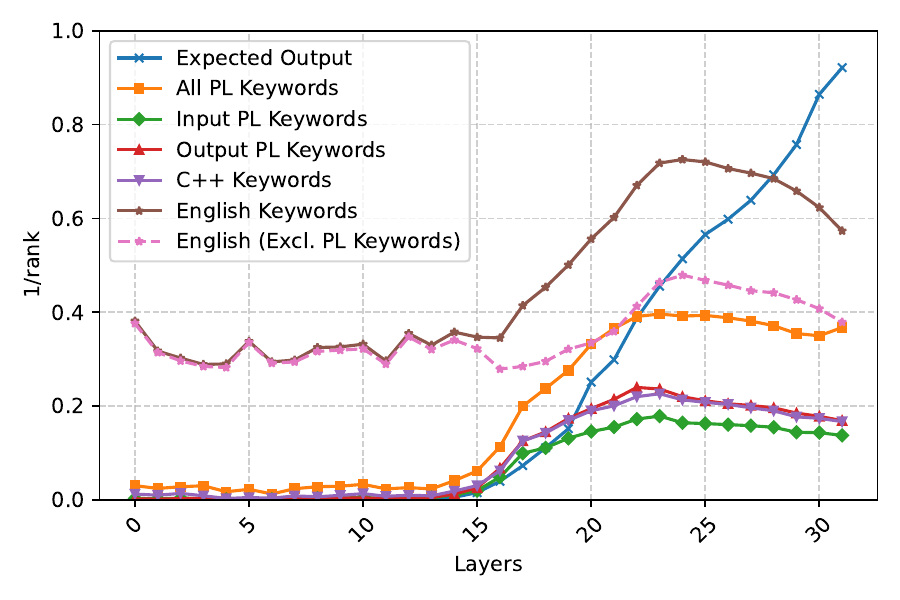}
        \caption{Llama 3.1 8B; $\frac{1}{rank}$ value.}
        \label{fig:llama31-rank}
    \end{subfigure}
    
    \caption{
    Language keyword probability or $\frac{1}{rank}$ value (best keyword rank) during translation task.
    The PLs contributing the most to each score, selected from the 22 PL keywords, are C++ and C\#.
    }
    \label{fig:rankprob}
\end{figure*}


\begin{figure}[t]
    \centering
    
    \begin{subfigure}{0.40\textwidth}
        \centering
        \includegraphics[width=\linewidth]{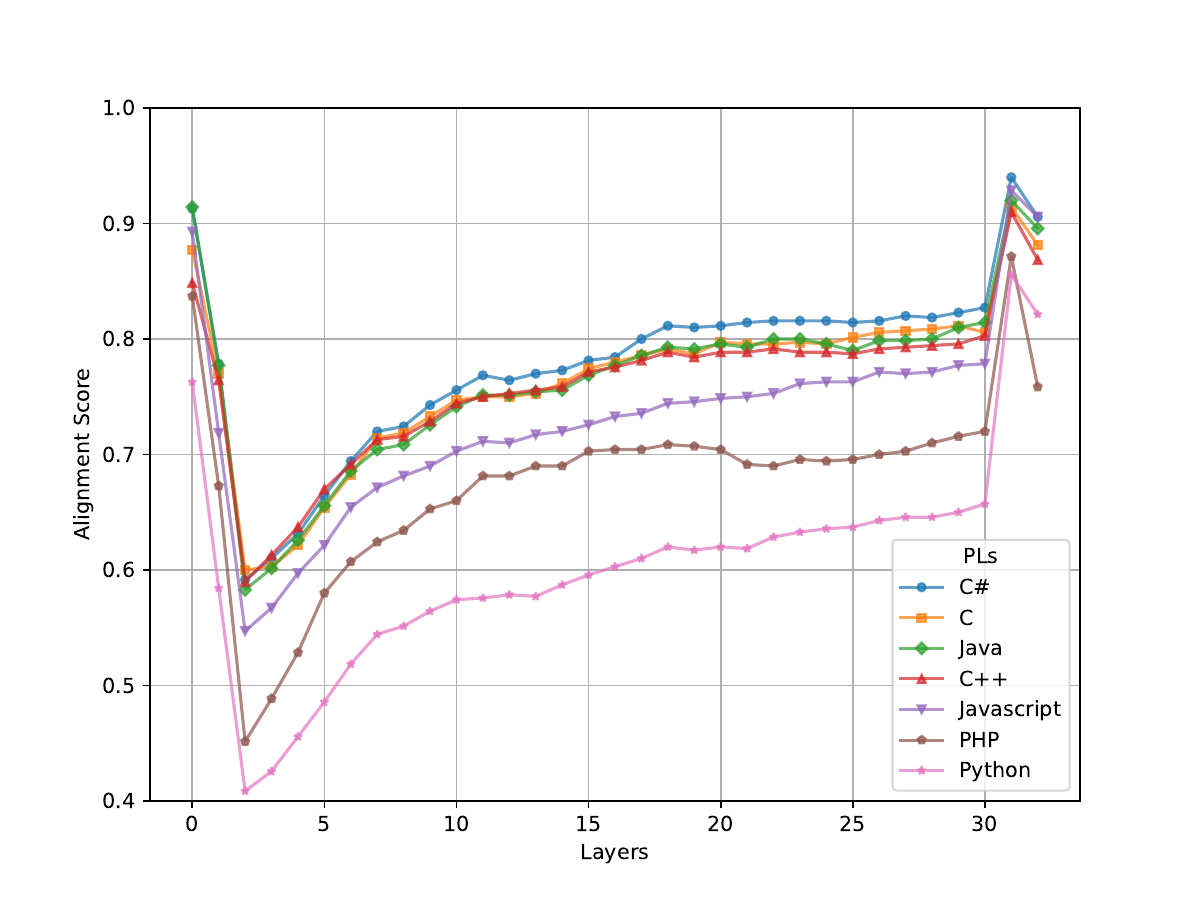}
        \caption{Code Llama 7B}
        \label{fig:codellama-mexa}
    \end{subfigure}
    \hfill 
    \begin{subfigure}{0.40\textwidth}
        \centering
        \includegraphics[width=\linewidth]{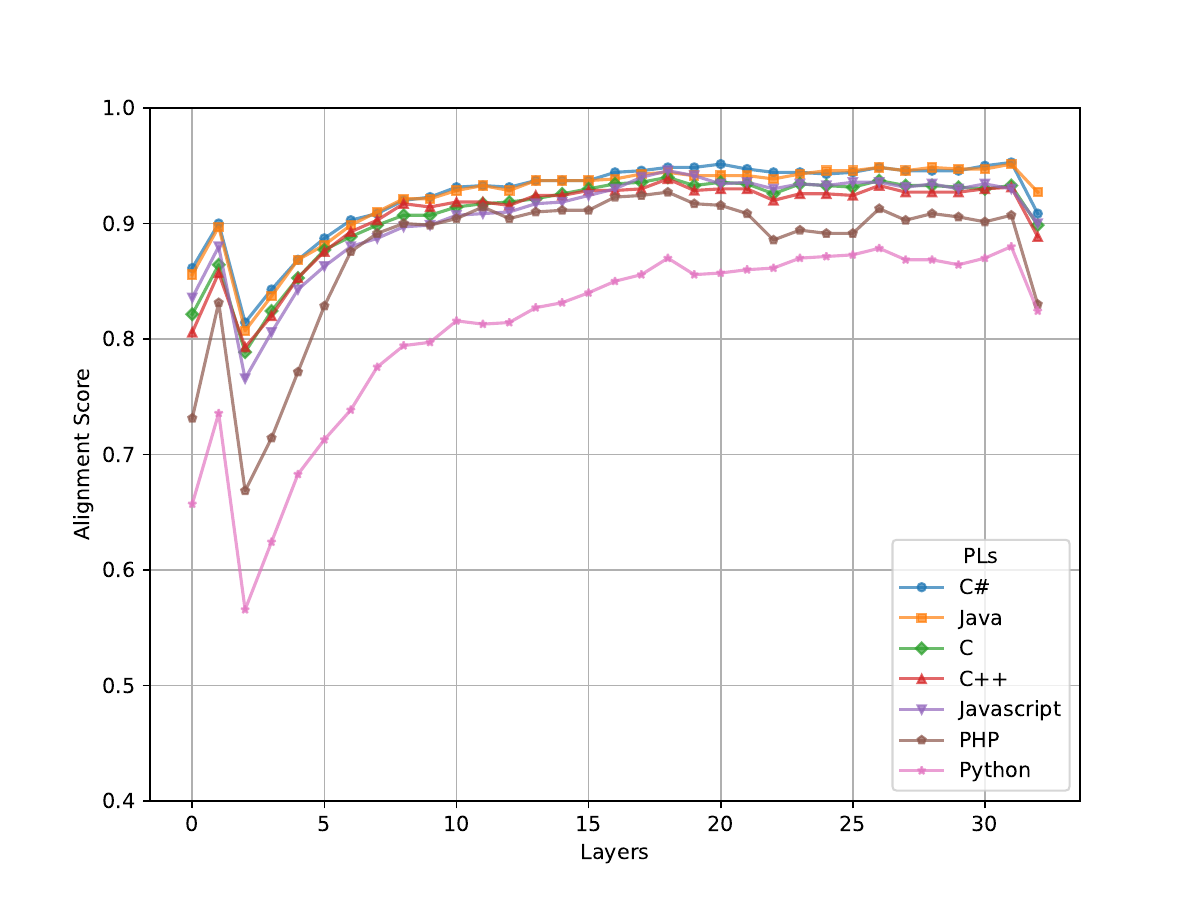}
        \caption{Llama 3.1 8B}
        \label{fig:llama31-mexa}
    \end{subfigure}
    
    \caption{
    MEXA alignment score. The minimum value of the MEXA alignment score is 0. The figures are limited to scores above 0.4 for better visualization.
    }
    \label{fig:mexa}
\end{figure}

\begin{figure}[t]
    \centering
    
    \begin{subfigure}{0.30\textwidth}
        \centering
        \includegraphics[width=\linewidth]{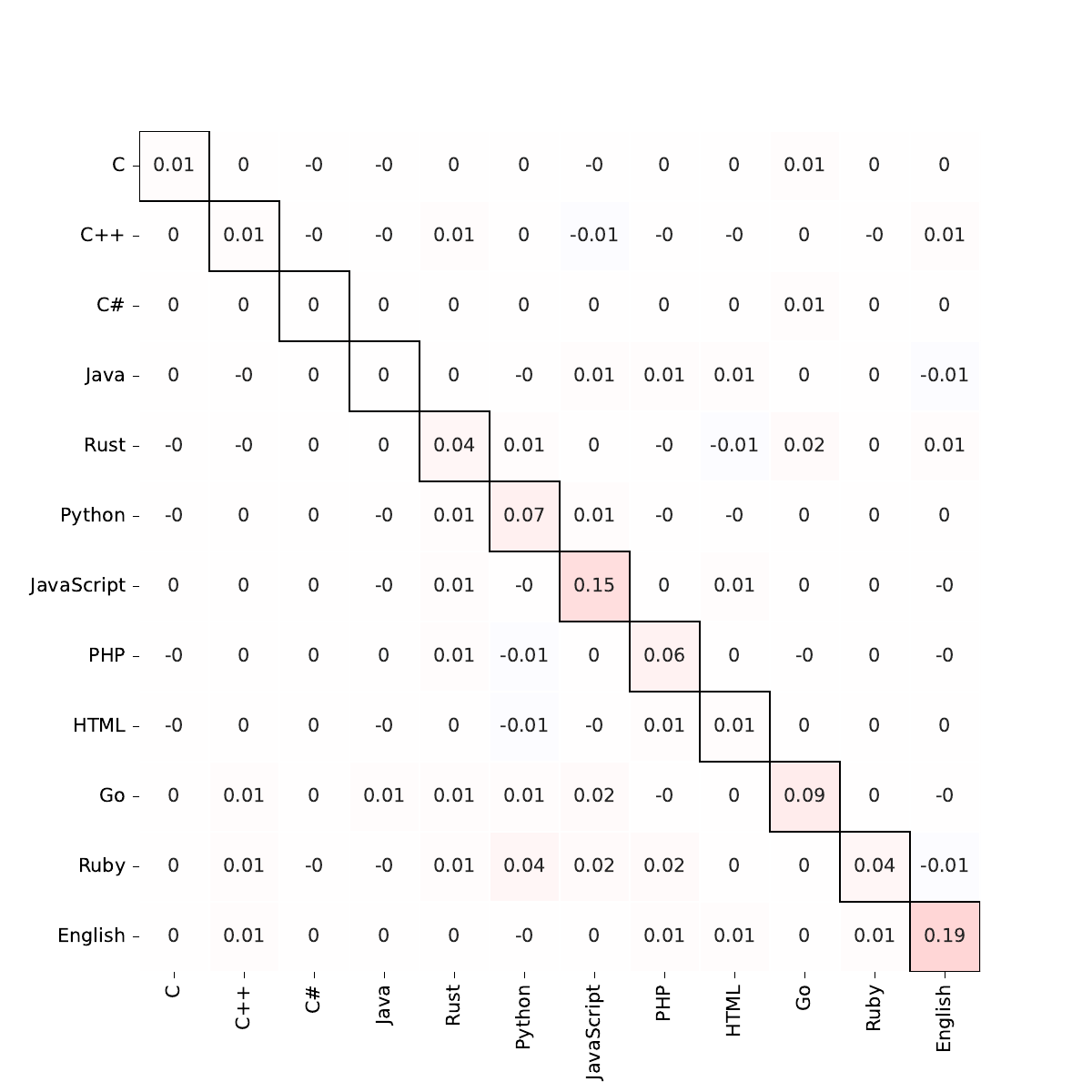}
        \caption{Code Llama 7B}
        \label{fig:codellama-pp}
    \end{subfigure}
    \begin{subfigure}{0.30\textwidth}
        \centering
        \includegraphics[width=\linewidth]{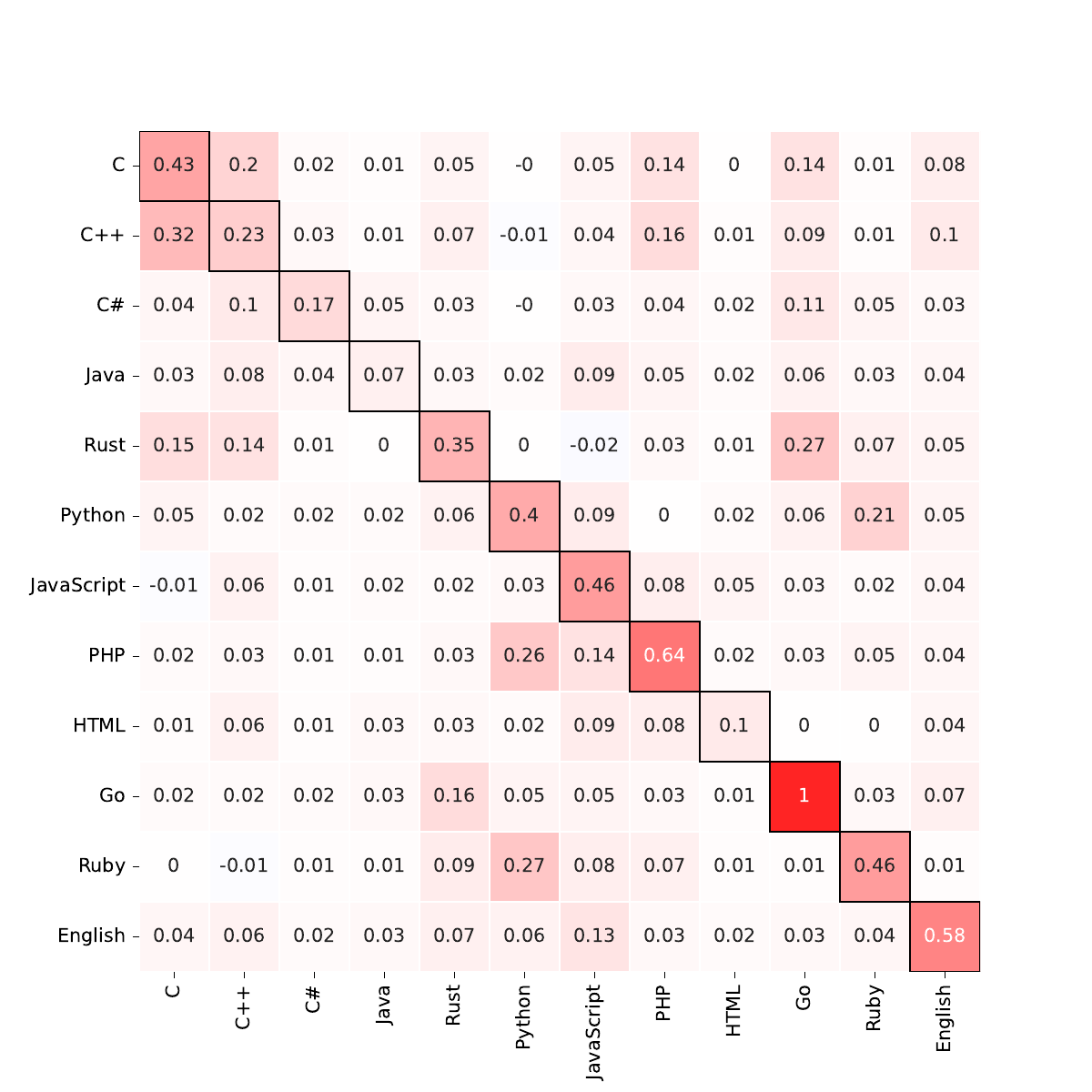}
        \caption{Llama 3.1 8B}
        \label{fig:llama31-pp}
    \end{subfigure}
    
    \caption{
    Impact of LAPE identification (\(\nu = 400, \tau = 0.95\)) on PPL increase. The element at row \(i\), column \(j\) represents the PPL change for language \(j\) due to perturbations in the language \(i\) region.
    }
    \label{fig:pp}
\end{figure}

\begin{figure*}[t]
    \centering
    
    \begin{subfigure}{0.45\textwidth}
        \centering
        \includegraphics[width=0.45\textwidth]{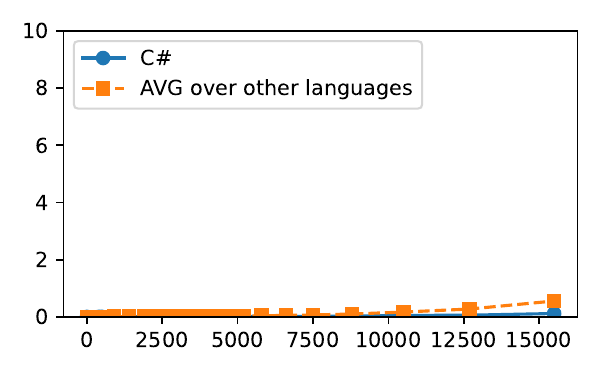}
        \includegraphics[width=0.45\textwidth]{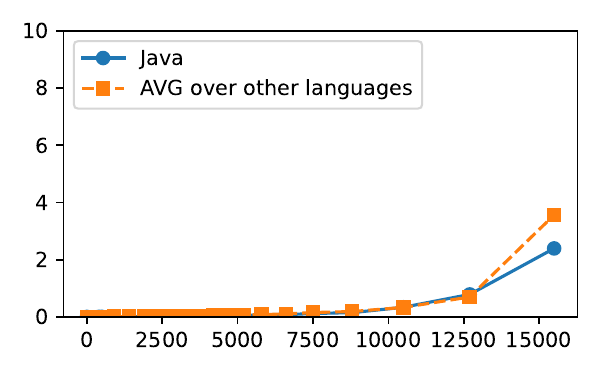}
        \includegraphics[width=0.45\textwidth]{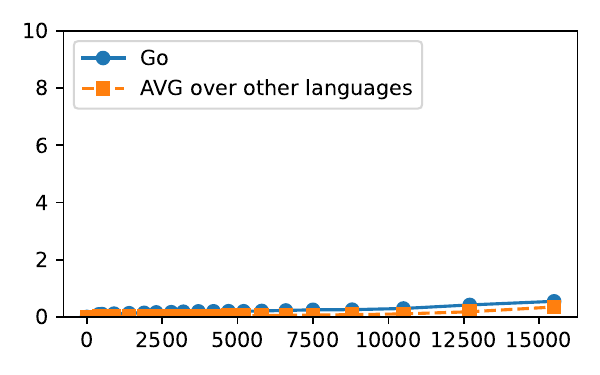}
        \includegraphics[width=0.45\textwidth]{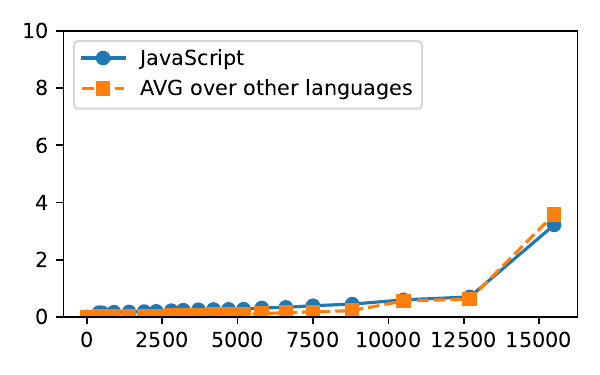}
        \includegraphics[width=0.45\textwidth]{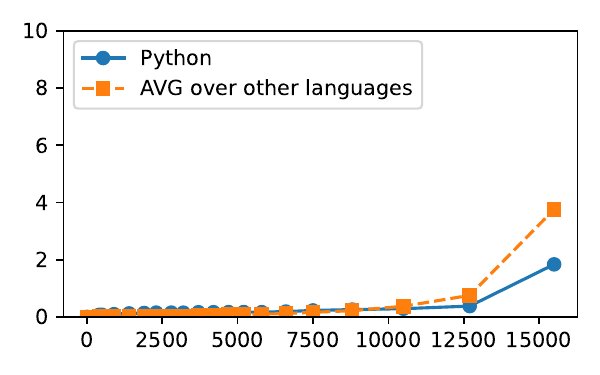}
        \includegraphics[width=0.45\textwidth]{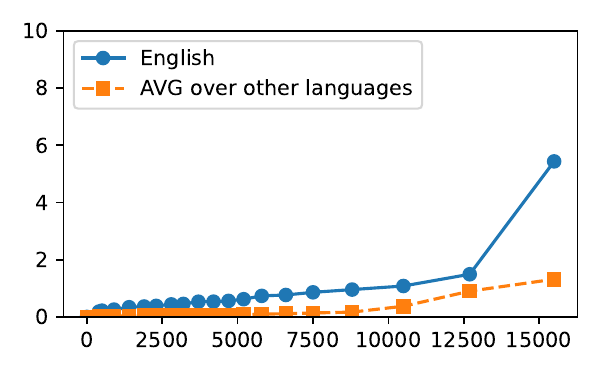}
        \caption{Code Llama 7B}
        \label{fig:ppchange-small-codellama}
    \end{subfigure}
    \hfill
    \begin{subfigure}{0.45\textwidth}
        \centering
        \includegraphics[width=0.45\textwidth]{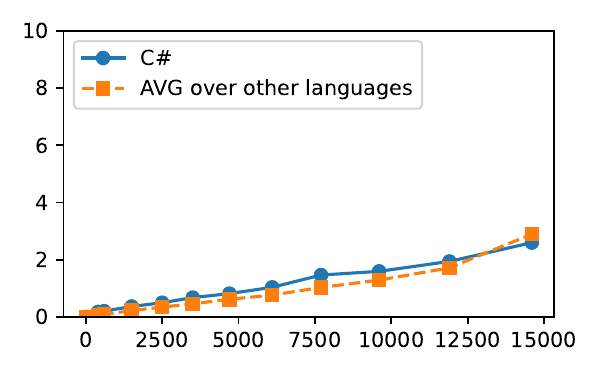}
        \includegraphics[width=0.45\textwidth]{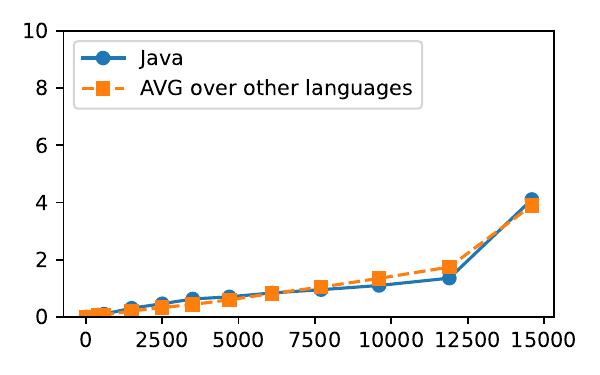}
        \includegraphics[width=0.45\textwidth]{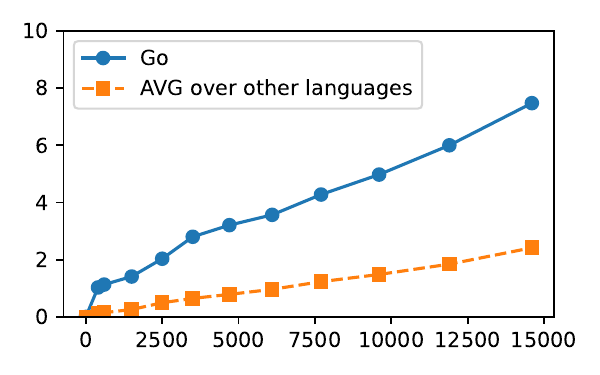}
        \includegraphics[width=0.45\textwidth]{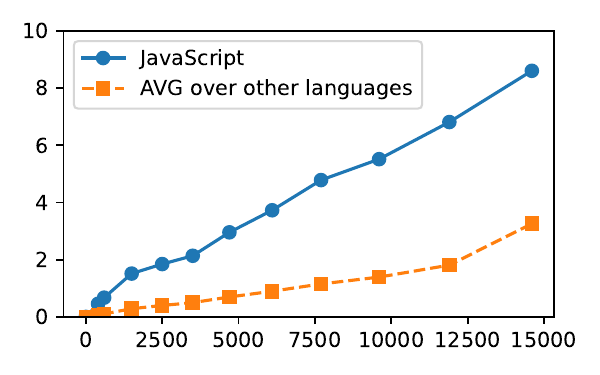}
        \includegraphics[width=0.45\textwidth]{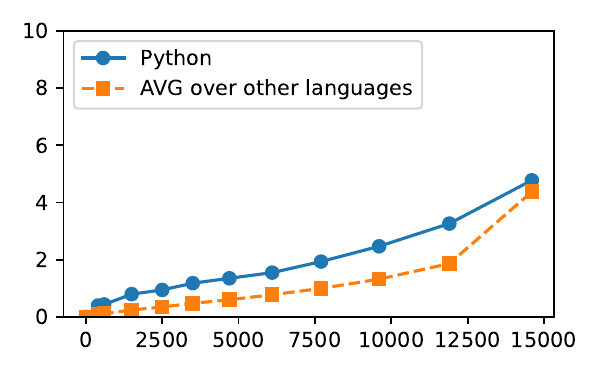}
        \includegraphics[width=0.45\textwidth]{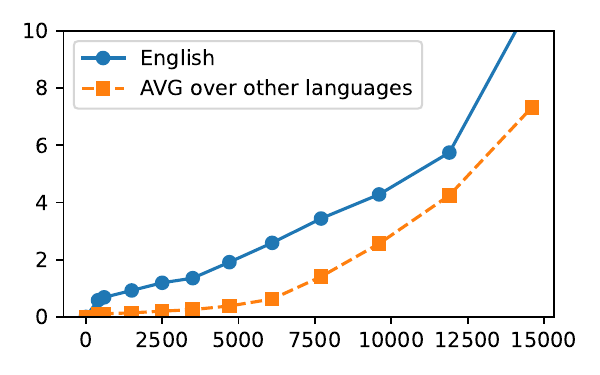}
        \caption{Llama 3.1 8B}
        \label{fig:ppchange-small-llama31}
    \end{subfigure}
    
    \caption{
    Impact of LAPE neuron identification. X-axis: Number of shared neurons for each language.
    Y-axis: Change in PPL across languages when deactivating the primary language’s neurons (e.g., English in the lower-right figure). \figref{ppchange-long} in \Appref{lape-more} shows the same figure for more languages.}
    \label{fig:ppchange-small}
\end{figure*}

\section{Results}

\subsection{Method 1: Interpreting latent embeddings}\seclabel{res-logitlens}

We present the results of interpreting latent embeddings for the translation task in \figref{rankprob}. Neither English nor PL keywords exhibit noticeable probability during the first half of the layers (Figures~\ref{fig:codellama-prob}, \ref{fig:llama31-prob}). Although these probabilities remain negligible, English keywords still appear among the top rank decoded tokens in the first half of the layers (Figures~\ref{fig:codellama-rank}, \ref{fig:llama31-rank}); this occurs much less frequently for PL keywords.

Around the half point (rougly, layer 15), the probabilities of English and PL keywords, as well as expected tokens, begin to rise sharply (Figures~\ref{fig:codellama-prob}, \ref{fig:llama31-prob}). English and PL keywords overtake the expected tokens at first. While expected token probability continues increasing until the final layers, English and PL keyword probabilities decline, particularly when English token probability crosses over the expected token probability. 

In the final layer, while the expected token holds rank = 1,
English keywords (excluding PL keywords) and PL keywords maintain a high and similar $\frac{1}{rank}$ value of $0.4$ each (Figures~\ref{fig:codellama-rank}, \ref{fig:llama31-rank}), indicating their presence among the top decoded tokens. Among individual PL keywords, the output PL dominates in both rank and probability measures, followed by popular PLs like C++ and C\# (which have some of the largest keyword sets), while the input PL has less influence. This distribution holds across different PL keywords: rising in the second half of the layers, peaking, and then decreasing in the final layers. Notably, many expected tokens are variable names, symbols, numbers, or punctuation, which typically fall outside the different PL keyword sets.

Regarding the comparison between CodeLlama 7B and Llama 3.1 8B: In terms of token probability, Llama 3.1 8B exhibits a slower initial rise in expected token probability, followed by a sharp increase in the top three layers. In contrast, CodeLlama 7B demonstrates a more gradual increase throughout. In terms of rank, CodeLlama 7B consistently shows a mainstream presence of English keywords. However, their distribution shifts: in the first half of the layers, they primarily consist of English keywords excluding PL keywords, while in the second half, they increasingly include English keywords that overlap with PL keywords. For Llama 3.1 8B, the presence of PL keywords also increases in the second half of the layers, reaching a $\frac{1}{rank}$ value of $0.7$ at layer index 23 for English keywords shared with PL keywords.

\subsection{Method 2: Cross-lingual alignment}\seclabel{res-mexa}

We present the results of cross-lingual alignment in \figref{mexa}. We compute alignment scores for all pairs of PLs and determine which PL aligns better with others. C\# achieves the best alignment overall across all layers in both models, though the difference between C-family PLs and Java is minimal. Both models show fewer alignments for Python. JavaScript is the best-aligned PL for both PHP and Python. The high alignment of C\# and C++ further supports the influence of popular PLs, as discussed in \Secref{res-logitlens}. This finding is also aligned with \citet{quan2025codeelo}, who find that although Python is the most familiar language for existing LLMs and competition-level code benchmarks, model performance improves over Python when responding in C++ for most of the models, including for the instruction-tuned version of the Llama 3.1 8B model.

The alignment scores consistently increase except for two instances: first, in the bottom layers (layer index 2 in Figures~\ref{fig:codellama-mexa}, \ref{fig:llama31-mexa}), where representations diverge from the ``input space''; and second, immediately before the final layer (layer index 31 in Figures~\ref{fig:codellama-mexa}, \ref{fig:llama31-mexa}), where they transition into the final ``output space''. The alignment of different PLs, especially in the layer preceding the final layer, indicates the high quality of the parallel data, as the alignment reaches values of $0.9$ in average.

Llama 3.1 8B achieves better MEXA alignment scores across all pairs compared to CodeLlama 7B.
This is not entirely unexpected: even though CodeLlama 7B and its instruction-tuned version are specifically trained for code, newer generic models of Llama, including Llama 3 8B~\citep{dubey2024llama} and its instruction-tuned version, achieve better scores in code generation tasks (as evaluated on LiveCodeBench~\citep{jain2024livecodebench}).\footnote{\href{https://huggingface.co/spaces/livecodebench/leaderboard}{\path{hf.co/spaces/livecodebench/leaderboard}}}

\begin{figure}[t]
    \centering

\includegraphics[width=0.85\linewidth]{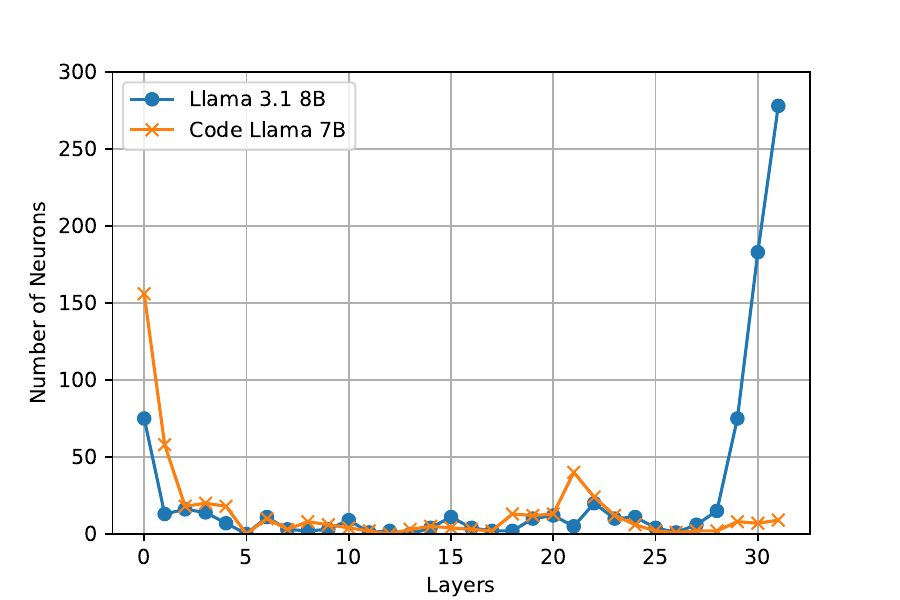}
\caption{Number of ``language-specific'' neurons, i.e., 
neurons that are exclusive to one PL and not shared with any
other PL, in the LAPE experiment with \(\nu = 1400, \tau = 0.95\).
The figure shows the total number of language-specific
neurons summed over all PLs.}
    \label{fig:exclusive}
\end{figure}

\subsection{Method 3: Language-specific neurons}\seclabel{res-lape}

We identify language-specific neurons for 11 PLs and English using LAPE. PPL change results ($\nu = 400, \tau = 0.95$) in \figref{pp} show that deactivating language-specific neurons has negligible effects on other languages, while more noticeably impacting the primary language—though this may not hold across all settings.
For other \(\nu\) values, we apply the LAPE neuron identification method and measure PPL changes by incrementally deactivating language-specific neurons for each primary language. The results for \(\tau = 0.95\) are shown in \figref{ppchange-small}, where we observe that LAPE fails to identify ``effective'' language-specific neurons for some languages. Effectiveness is indicated by a larger PPL change for the primary language compared to other languages when deactivating the primary language-specific neurons.

In most cases, increasing the number of neurons enlarges the PPL gap between primary and other languages in Llama 3.1. However, for C\# and Java (and C++ and HTML in \figref{ppchange-long}) the gap is less pronounced. Interestingly, C\# and Java are the PLs with the highest alignment in Llama 3.1, as shown in \Secref{res-mexa}. Additionally, C\# has one of the largest PL keyword sets appearing frequently in intermediate layers, as shown in \Secref{res-logitlens}. This suggests that the identified specific neurons for these languages are more shared across languages. In other words, for PLs such as C\# and Java, which closely align with multiple other PLs, distinguishing language-specific neurons is more challenging.

For CodeLlama 7B, even though some effective language-specific neurons exist, the PPL change is not significant (Figures~\ref{fig:codellama-pp}, \ref{fig:ppchange-small-codellama}). When the number of deactivated neurons exceeds 12,500, the impact on other languages sometimes surpasses that on the primary language. The only language for which CodeLlama 7B identifies effective specific neurons with a larger PPL change margin is English. This suggests that PL neurons in CodeLlama 7B are highly shared, possibly due to its training recipe, where the pre-training phase following Llama 2 initialization primarily focuses on code.  

To further investigate what makes language-specific neurons effective in Llama 3.1 8B but not in CodeLlama 7B, we examine the language-specific neurons selected for all PLs that are ``exclusive'' to each PL, as shown in \figref{exclusive} for \(\nu = 1400, \tau = 0.95\). Other \(\nu\) settings exhibit a similar distribution. In general, most language-specific neurons in both models and across most languages are selected from the bottom layers (indices 0 to 4), followed by layer indices 18 to 22 in both models. However, those that are exclusive to a specific PL are predominantly selected from the top layers (indices 29 to 31). Notably, LAPE selects more exclusive neurons from top layers for Llama 3.1 8B than CodeLlama 7B as shown in \figref{exclusive}. This aligns with the fact that top layers serve for token generation, where the LLM must handle the ``output space'' and map it to the expected token. If exclusive language-specific neurons exist for each primary language at the top layers, deactivating that language's neurons only affects the PPL of that language. However, if there are no such exclusive neurons, it affects PPL of other languages as well.

\section{Discussion and Implications}

Our findings suggest several strategies for building more efficient multilingual code models.

1) Since English and certain PLs are centrally located in the model’s concept space, these could serve as intermediate representations for multilingual code translation, minimizing the distance between source and target languages.

2) The distribution of neuron types across layers -- shared/general in bottom layers, specific in top -- supports modular architectures where base layers encode general syntax/semantics and top layers can be swapped or specialized for specific languages.

3) For closely aligned PLs (e.g., Java and C\#), shared representations could enable parameter sharing or adapter-based methods for lightweight multilingual support, while only tuning minimal additional weights.

4) Some languages enforce object-oriented programming, while others support it optionally. This structural difference may lead the model to develop stronger internal representations for languages with stricter paradigms, potentially introducing some bias in code generation. Other differences in language design and idiomatic usage can influence the model’s behavior when generating code across languages. Recognizing these factors could help improve the generalization capabilities of multilingual code models.

\section{Related work}

\textbf{Pivot language.}
\citet{wendler2024llamas} use logit lens~\citep{nostalgebraist2020interpreting} to show that English acts as a kind of ``pivot'' language in English-centric LLMs, such as Llama-2~\citep{touvron2023llama2}, enabling these models to solve complex semantic tasks in a non-English language by detouring through English internal states before generating non-English text.
%
Building on this idea, \citet{wu2024semantic} propose the semantic hub hypothesis, which suggests that the same phenomenon could occur not only across different languages but also across different modalities. As one of these modalities, they introduce code. Their analysis focuses solely on Python within the Llama 2 model. Since obtaining semantically equivalent English-Python  pairs is challenging, they test only a few targeted cases, such as the English token ``and'' and its Python counterpart ``,''. Using logit lens, they show that in the intermediate layers, the expected Python token is closer to ``and'' than to other tokens such as ``or'' and ``not.''  
In our work, we focus exclusively on PLs and consider seven PLs. As noted by \citet{wu2024semantic}, obtaining semantically equivalent English-PL pairs is challenging. Instead, we analyze keyword sets—comprising keywords from 22 PLs and an English dictionary—through a translation task across 42 directions. This allows us to examine which PLs and English-derived tokens appear in the model’s intermediate layers and are closer to its concept space, both in terms of probability and rank. Our findings reveal that not only English but also other PLs contribute to the model's concept space.

\textbf{Neuron-level interpretability.}
Initially, language-specific components were studied in  neural machine translation  using small language models~\citep{lin-etal-2021-learning, xie-etal-2021-importance, zhang2021share}. Later, the role of FFNs within LLMs was explored in several studies, highlighting their function as key-value memories for storing factual and linguistic knowledge~\citep{geva-etal-2021-transformer, geva-etal-2022-transformer, ferrando-etal-2023-explaining}. However, these analyses typically investigate neuron behavior, focusing on monolingual settings in natural languages and PLs.
Building on methods explored in investigations on the role of FFNs within LLMs and considering clear evidence that LLMs exhibit significant overlap in their embeddings across languages—particularly among those from the same linguistic family~\citep{doddapaneni2021primer}—several recent studies~\citep{xie-etal-2021-importance, tang-etal-2024-language, zhao2024how, kojima-etal-2024-multilingual, wang2024sharing, bhattacharya-bojar-2023-unveiling, bhattacharya2024understanding, mueller-etal-2022-causal, liu-etal-2024-unraveling, dumas2024how, liu2025relation} have investigated the existence of language-specific neurons and internal mechanisms for natural languages, especially within the FFN layers of LLMs.
Just as there are many natural languages, there are also many PLs. However, no research has explored the existence of language-specific neurons for PLs, even though LLMs are typically pre-trained on a mixture of these languages. Building on this, our work adopts the method proposed by \citet{tang-etal-2024-language} to identify PL-specific neurons. This approach enables a scalable and targeted analysis of neurons for many PLs using only raw PL data.

\textbf{Interpretability for code.}
Interpretability in language models for code-related tasks remains under-explored, with most research focusing on attention layers~\citep{mohammadkhani2023explaining, wan2022they, paltenghi2021thinking, liu2024reliability}. Our work is closest to \citet{haider2024looking}, who analyze FFN layers. They analyze two GPT-based models~\citep{xu2022systematic, nijkamp2022codegen} for three PLs, showing that lower layers capture syntax while higher layers encode abstract concepts and semantics. They demonstrate that concepts are stored in the FFN layers and can be edited without compromising code language model performance. However, their analysis is performed in monolingual settings, while our work investigates the relationship between PLs to determine if they share concepts and neurons in coding tasks.

\section{Conclusion}

In this study, we investigate how LLMs represent programming languages (PLs) in their concept space using the logit lens method. We observe that English and PL keywords appear in intermediate layers, with notable probabilities in the latter half. Initially, these keywords surpass the expected output tokens, but as the probabilities of expected tokens increase and overtake those of English and PL keywords, the probabilities of the latter decline. We further investigate the existence of language-specific neurons using the language activation probability entropy (LAPE) method. Our analysis reveals that language-specific neurons can be identified for most languages in the Llama 3.1 model, but not for PLs such as Java and C\#, which align closely with other PLs. We find that language-specific neurons are concentrated in the bottom layers, while neurons exclusive to each PL are located in the top layers. These findings deepen our understanding of LLMs' inner workings in the context of PLs and provide valuable insights for interpretability in code-related tasks.

\section*{Limitations}
We are aware of three main limitations of our work.

First, parts of our analysis rely on a super-parallel dataset, which is limited to seven languages due to source constraints. To our knowledge, no super-parallel dataset with a broader language set is publicly available. A potential solution is to generate super-parallel data for more languages using more powerful LLMs and validate it through unit tests to ensure quality and consistency.  

Second, while we use keywords to interpret latent embeddings, a more precise approach would involve constructing a dictionary mapping PL keywords to each other and their English equivalents. However, this is not always feasible, as some PL keywords lack direct English meanings or map to multiple tokens.  

Third, we hypothesize that the ineffectiveness of neuron identification for CodeLlama 7B stems from its training recipe, but further investigation across other models could be beneficial. Our analysis focuses on PLs in Llama-based architectures, which underlie many state-of-the-art models, but it's important to explore other architectures for broader validation.

\section*{Acknowlegments}
This research was supported by DFG (grant SCHU 2246/14-1).

\bibliography{main}

\appendix

\section{Impact of LAPE neuron identification}\applabel{lape-more}

\begin{figure*}[t]
    \centering
    
    \begin{subfigure}{0.45\textwidth}
        \centering
        \includegraphics[width=0.45\textwidth]{figures/codellama-ppchange/Csharpneurons.pdf}
        \includegraphics[width=0.45\textwidth]{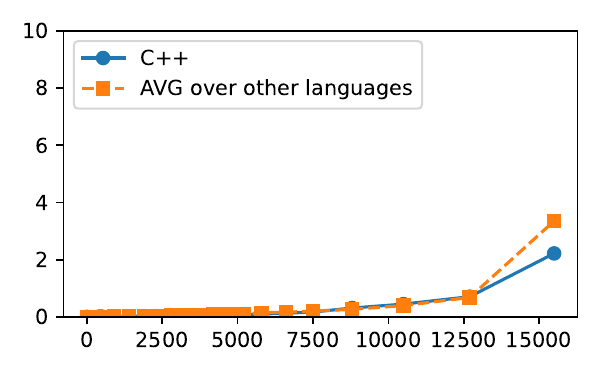}
        \includegraphics[width=0.45\textwidth]{figures/codellama-ppchange/Javaneurons.pdf}
        \includegraphics[width=0.45\textwidth]{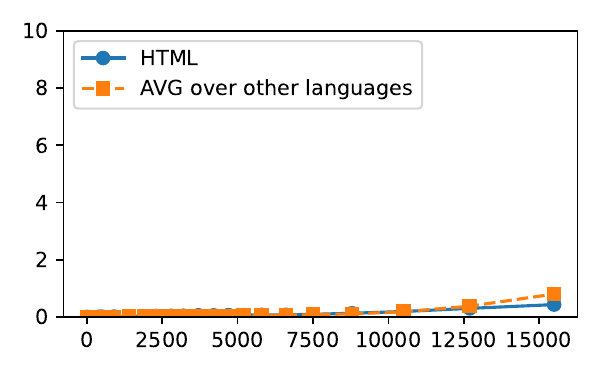}
        \includegraphics[width=0.45\textwidth]{figures/codellama-ppchange/Goneurons.pdf}
        \includegraphics[width=0.45\textwidth]{figures/codellama-ppchange/JavaScriptneurons.pdf}
        \includegraphics[width=0.45\textwidth]{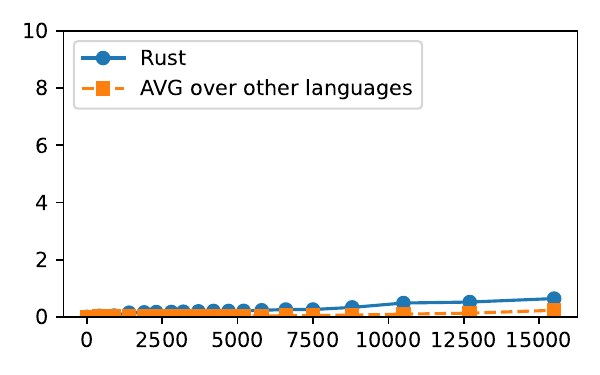}
        \includegraphics[width=0.45\textwidth]{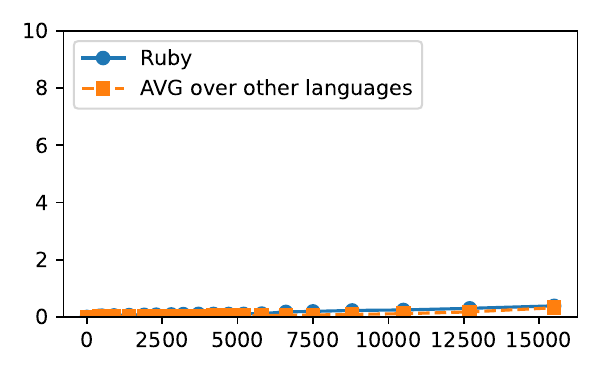}
        \includegraphics[width=0.45\textwidth]{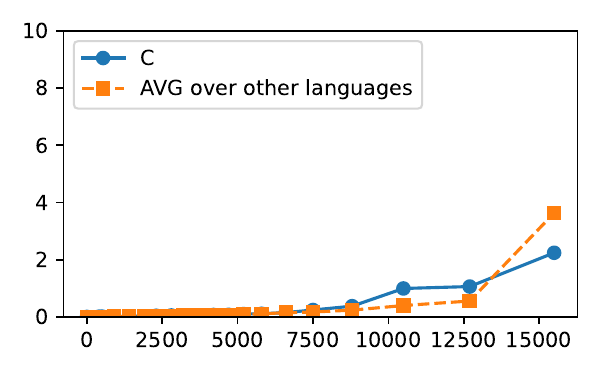}
        \includegraphics[width=0.45\textwidth]{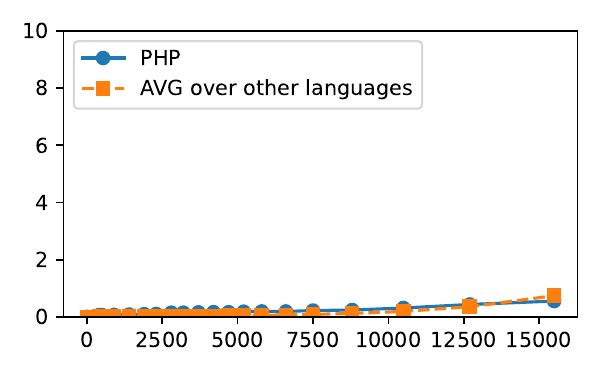}
        \includegraphics[width=0.45\textwidth]{figures/codellama-ppchange/Pythonneurons.pdf}
        \includegraphics[width=0.45\textwidth]{figures/codellama-ppchange/Englishneurons.pdf}
        \caption{Code Llama 7B}
        \label{fig:ppchange-long-codellama}
    \end{subfigure}
    \hfill
    \begin{subfigure}{0.45\textwidth}
        \centering
        \includegraphics[width=0.45\textwidth]{figures/llama31-ppchange/Csharpneurons.pdf}
        \includegraphics[width=0.45\textwidth]{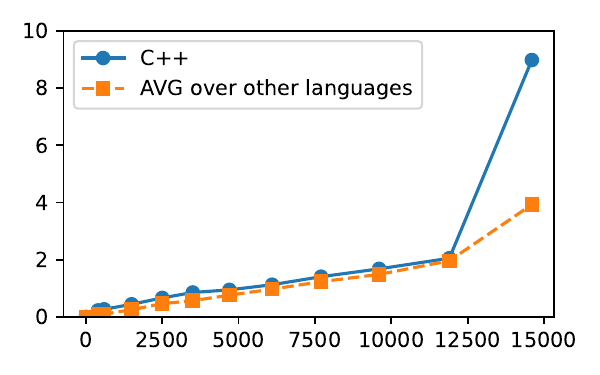}
        \includegraphics[width=0.45\textwidth]{figures/llama31-ppchange/Javaneurons.pdf}
        \includegraphics[width=0.45\textwidth]{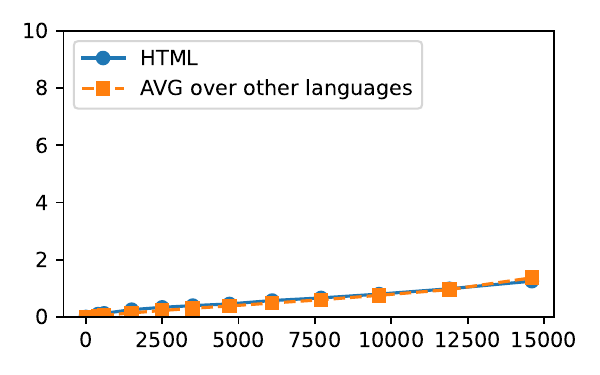}
        \includegraphics[width=0.45\textwidth]{figures/llama31-ppchange/Goneurons.pdf}
        \includegraphics[width=0.45\textwidth]{figures/llama31-ppchange/JavaScriptneurons.pdf}
        \includegraphics[width=0.45\textwidth]{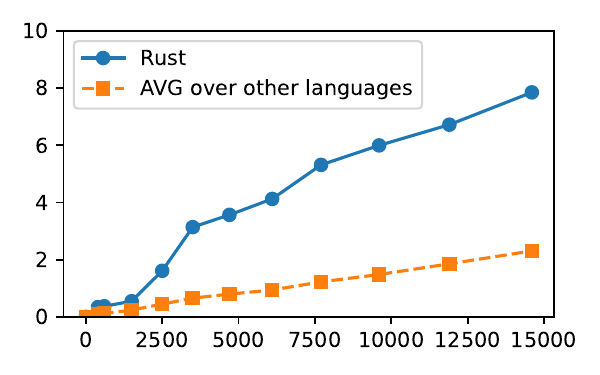}
        \includegraphics[width=0.45\textwidth]{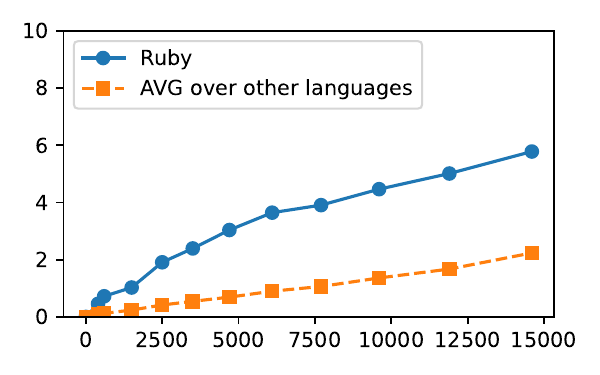}
        \includegraphics[width=0.45\textwidth]{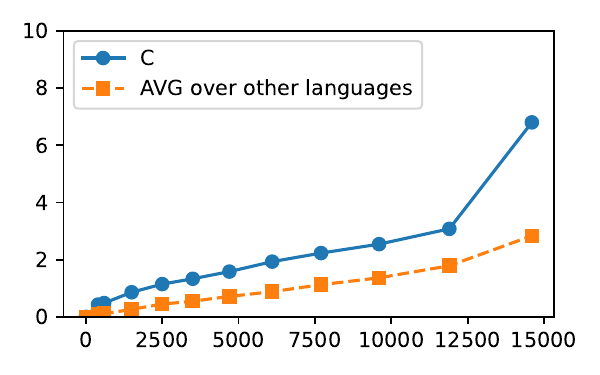}
        \includegraphics[width=0.45\textwidth]{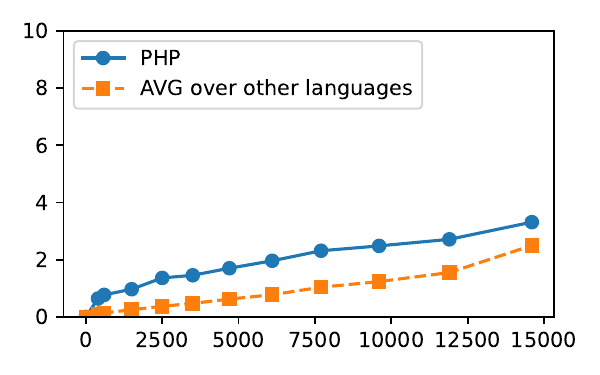}
        \includegraphics[width=0.45\textwidth]{figures/llama31-ppchange/Pythonneurons.pdf}
        \includegraphics[width=0.45\textwidth]{figures/llama31-ppchange/Englishneurons.pdf}
        \caption{Llama 3.1 8B}
        \label{fig:ppchange-long-llama31}
    \end{subfigure}
    
    \caption{
    Impact of LAPE neuron identification. X-axis: Number of shared neurons for each language.
    Y-axis: Change in PPL across languages when deactivating the primary language’s neurons (e.g., English in the lower-right figure).}
    \label{fig:ppchange-long}
\end{figure*}

We show the complete version of \figref{ppchange-small} in \figref{ppchange-long}, covering more PLs.

\end{document}